%% file: main.tex
\documentclass[10pt,twocolumn,letterpaper]{article}

\usepackage[pagenumbers]{cvpr} %

\input{abbreviation}

\input{preamble}

\definecolor{cvprblue}{rgb}{0.21,0.49,0.74}
\usepackage[pagebackref,breaklinks,colorlinks,citecolor=cvprblue]{hyperref}

\def\paperID{*****} %
\def\confName{3DV\xspace}
\def\confYear{2026\xspace}

\title{Finding NeMO\@: A Geometry-Aware Representation of Template Views for Few-Shot Perception}

\author{
Sebastian Jung \quad
Leonard Klüpfel \quad
Rudolph Triebel \quad
Maximilian Durner\\[2mm]
German Aerospace Center (DLR) \\
{\tt\small \{Sebastian.Jung, Leonard.Kluepfel, Rudolph.Triebel, Maximilian.Durner\}@dlr.de}
}
\begin{document}
\twocolumn[{
	\maketitle
	\begin{center}
		\includegraphics[width=0.85\linewidth]{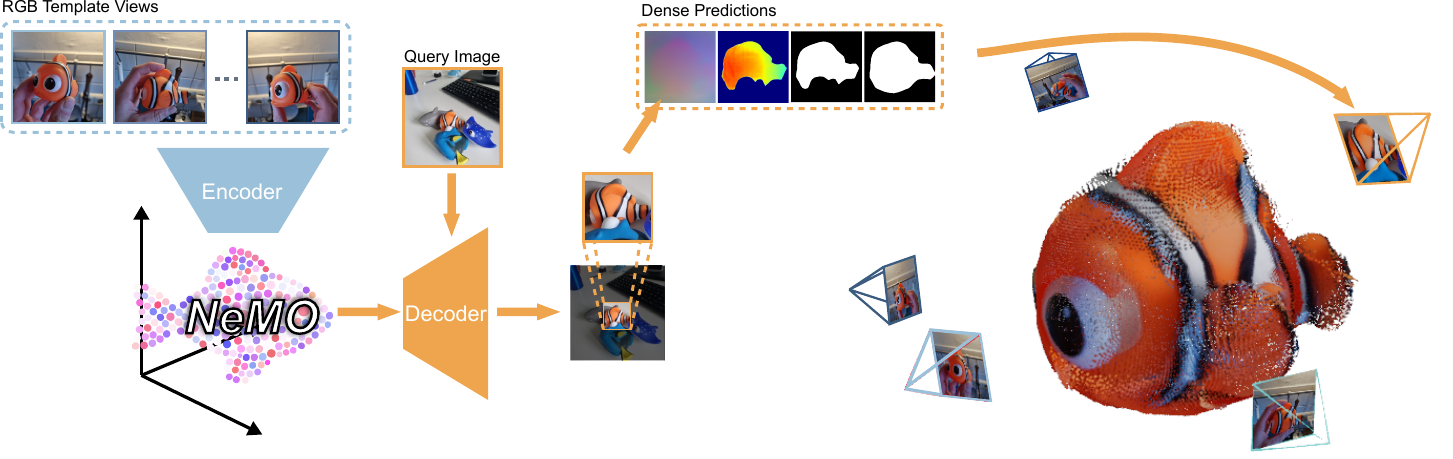}
	\end{center}
	\vspace{-0.5cm}
	\captionsetup{type=figure}
	\captionof{figure}{%
		\textbf{Overview.}
		Our method uses a multi-view encoder to generate an object-centric geometric encoding called \textit{Neural Memory Object (NeMO)} with its own coordinate system from a set of RGB images depicting an object unseen during training. A decoder uses the NeMO to retrieve dense predictions allowing us to detect, segment, estimate the objects surface and determine the camera-to-object position on an RGB query image.
		Even in cluttered scenes, our method is able to find the object, which we can use to crop the corresponding region of interest, demonstrating that our method can be used for multi-stage perception pipelines. Images were captured using a normal smartphone.
	}\label{fig:cover_fig}
	\vspace{0.3cm}
}]
\input{0_abstract}
\input{1_introduction}
\input{2_related_work}
\input{3_method}
\input{4_experiment}

\input{5_conclusion}

{
    \small
    \bibliographystyle{ieeenat_fullname}
    \bibliography{main.bib}
}
\input{6_supplementary}

\end{document}

%% file: abbreviation.tex
\usepackage[acronym]{glossaries}

\newacronym{AP}{AP}{Average Precision}
\newacronym{AR}{AR}{Average Recall}
\newacronym{BOP}{BOP}{Benchmark for 6D Object Pose Estimation}
\newacronym{CNN}{CNN}{Convolutional Neural Network}

\newacronym{DL}{DL}{Deep Learning}

\newacronym[plural=GPUs,firstplural=graphics processing units (GPUs)]{GPU}{GPU}{graphics processing unit}

\newacronym{IoU}{IoU}{Intersection over Union}
\newacronym[plural=ICPs]{ICP}{ICP}{Interative Closest Point}
\newacronym{mAP}{mAP}{Mean Average Precision}
\newacronym{MLP}{MLP}{Multi-Layer Perceptron}
\newacronym{MSPD}{MSPD}{Maximum Symmetry-Aware Projection Distance}
\newacronym{MSSD}{MSSD}{Maximum Symmetry-Aware Surface Distance}
\newacronym[plural=NeMOs,firstplural=Neural Memory Objects (NeMOs)]{NeMO}{NeMO}{Neural Memory Object}
\newacronym[plural=NERFs,firstplural=Neural Radiance Fields (NERFs)]{NERF}{NERF}{Neural Radiance Field}

\newacronym{PnP}{PnP}{Perspective-n-Point}
\newacronym{pca}{PCA}{Principal Component Analysis}
\newacronym{PBR}{PBR}{Physically-Based Rendering}
\newacronym{RoI}{RoI}{Region of Interest}
\newacronym{RANSAC}{RANSAC}{Random Sample Consensus}
\newacronym{SfM}{SfM}{Structur-from-Motion}
\newacronym{SDF}{SDF}{Signed Distance Function}
\newacronym[plural=UDFs,firstplural=Unsigned Distance Fields (UDFs)]{UDF}{UDF}{Unsigned Distance Field}

\newacronym[plural=ViTs,firstplural=Vision Transformers (ViTs)]{ViT}{ViT}{Vision Transformer}
\newacronym{VSD}{VSD}{Visible Surface Discrepancy}

%% file: preamble.tex
\usepackage[dvipsnames]{xcolor}
\usepackage{algorithm}
\usepackage{algpseudocode}
\usepackage{pifont}
\usepackage{caption}

\newcommand{\todo}[1]{{\color{red}#1}}
\newcommand{\todoLater}[1]{{\color{red}#1}}
\newcommand{\TODO}[1]{\textbf{\color{red}[TODO: #1]}}
\renewcommand{\TODO}[1]{}
\renewcommand{\todo}[1]{}
\renewcommand{\todoLater}[1]{}

\newcommand{\suppref}[1]{Supplementary~\cref{#1}}

%% file: 0_abstract.tex
\begin{abstract}
    We present \acrfull{NeMO}, a novel object-centric representation that can be used to detect, segment and estimate the 6DoF pose of objects unseen during training using RGB images. 
    Our method consists of an encoder that requires only a few RGB template views depicting an object to generate a sparse object-like point cloud using a learned \acrshort{UDF} containing semantic and geometric information.
    Next, a decoder takes the object encoding together with a query image to generate a variety of dense predictions. 
    Through extensive experiments, we show that our method can be used for few-shot object perception without requiring any camera-specific parameters or retraining on target data.
    Our proposed concept of outsourcing object information in a \acrshort{NeMO} and using a single network for multiple perception tasks enhances interaction with novel objects, improving scalability and efficiency by enabling quick object onboarding without retraining or extensive pre-processing.
    We report competitive and state-of-the-art results on various datasets and perception tasks of the \acrshort{BOP} benchmark, demonstrating the versatility of our approach.
    \url{https://github.com/DLR-RM/nemo}
    \vspace*{12pt}
\end{abstract}

%% file: 1_introduction.tex
\vspace*{-3em}
\section{Introduction}\label{sec:intro}
Objects play a central role in our daily lives, and recognizing them in images is critical for applications such as robotics, augmented reality, and autonomous systems.
Recent advances in deep learning and computer vision have greatly improved object perception, especially in \textit{model-based} approaches that leverage 3D CAD models to train deep networks for detection, segmentation, and pose estimation~\cite{hodan_bop_2024}. These methods benefit from large-scale synthetic training~\cite{Denninger2023} and can specialize in specific objects or categories, achieving impressive performance when 3D models are available at test time. However, in many real-world scenarios, it is impractical to assume access to a 3D model for every object. As a result, \textit{model-free perception} has become a growing research focus, aiming to rapidly onboard and recognize novel objects.

When a CAD model is unavailable during training but provided at inference, recent methods adapt perception models using rendered templates of the target object, leveraging the geometric and textural cues from these synthetic views.
Some approaches fine-tune object-specific networks in minimal time~\cite{bop_frtpose}, while others use general-purpose features to recognize new objects without retraining~\cite{nguyen_cnos_2023}.
A common strategy compares features between template views and query images, followed by post-processing~\cite{nguyen_gigapose_2024, labbe_megapose_2023}. However, these approaches scale poorly with the number of templates and rely on pairwise local comparisons~\cite{sarlin_superglue_2020, sun_loftr_2021, detone_superpoint_2018}, without jointly reasoning over all views.
While still affected by the sim-to-real gap, such methods enable fast onboarding of CAD models.

In model-free object perception -- where no CAD model is available at any stage -- the challenge is greater. Synthetic training tailored to the object is infeasible, and geometric information must be extracted from real reference images, where object-to-camera poses are typically unknown. Some methods rely on template or local feature matching using real images~\cite{lin_relpose_2024, he_fs6d_2022}, while others employ neural fields~\cite{mildenhall2020nerfrepresentingscenesneural} to reconstruct CAD-like geometry~\cite{wen_bundlesdf_2023, wen_foundationpose_2024}, requiring extrinsic pose information.

To overcome the limitations of existing perception systems in terms of generalization, scalability, and efficiency, we propose a novel encoder-decoder architecture trained on a large-scale synthetic dataset.
Our method constructs a geometry-aware representation, termed \gls{NeMO}, from a set of unordered, object-centric RGB images, without requiring camera calibration or pose annotations.
\gls{NeMO} is formulated as a sparse, continuous point cloud that encapsulates both semantic and geometric features observed from multiple viewpoints.
Unlike conventional encoder-decoder models that compress inputs into a single latent vector~\cite{Sundermeyer_2018_ECCV,Shi24arxiv-crisp}, \gls{NeMO} preserves a structured, interpretable abstraction of object geometry, enabling transformations such as translation, rotation, and scaling without reliance on a 3D CAD model.
Critically, the object-specific information is disentangled from the decoder's parameters, enabling object-agnostic inference and robust generalization to novel instances not seen during training.
The point-based representation supports incremental refinement through the addition of new views, without necessitating reprocessing of previous observations.
Furthermore, the decoupling of encoding and decoding facilitates efficient deployment, as NeMO can be precomputed offline, rendering inference time invariant to the number of input views and improving scalability for real-world applications.

We extensively ablate the representation to analyze its potential and quantitatively demonstrate its performance on multiple object perception tasks, achieving competitive and state-of-the-art results on model-free and model-based unseen object benchmarks. Additionally, we qualitatively show its potential for object surface reconstruction.
Concretely, our contributions are threefold:
\begin{itemize}
  \item We propose the \textbf{\gls{NeMO}}, a geometry-sensitive, and compact representation of template views well-suited for few-shot object perception tasks.
  \item We evaluate our encoder-decoder network on the task of model-based and -free \textbf{few-shot detection, segmentation and pose estimation of unseen objects}.
  Without training on test objects, we perform competitively and partially better against state-of-the-art methods.
  \item We contribute an object-centric \textbf{diverse synthetic dataset}, mimicking realistic cluttered scenes that balances the occurrences of different object classes to advance research on related topics.
\end{itemize}

%% file: 2_related_work.tex
\begin{figure*}[t]
    \centering
    \includegraphics[width=0.95\linewidth]{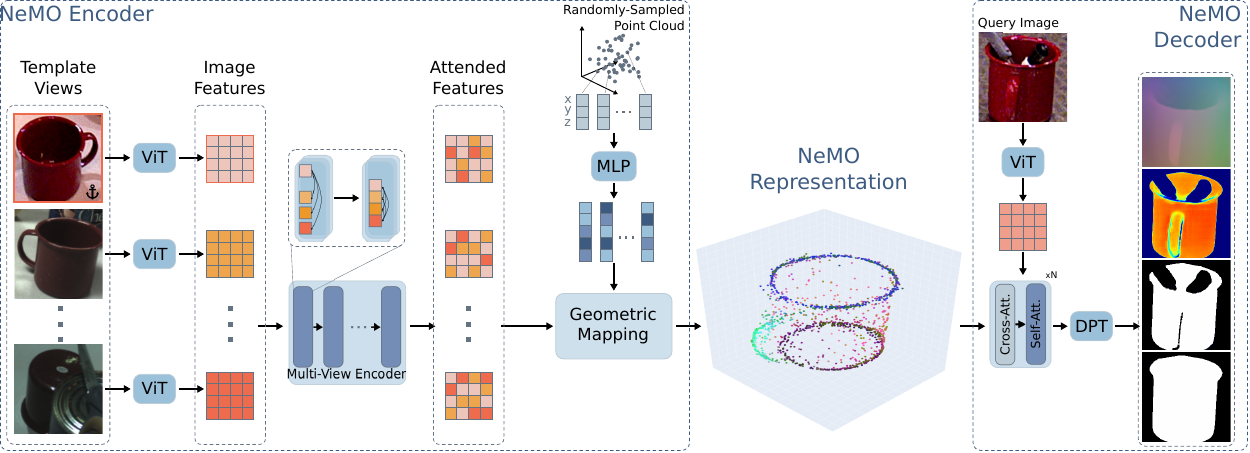}
    \caption{\textbf{Overview of the \gls{NeMO} approach.} RGB template views are first processed by a \acrshort{ViT}~\cite{dosovitskiy2020image} and a multi-view encoder~\cite{jiang_leap_2023}, producing updated image features. To incorporate spatial information, a randomly-sampled point cloud is processed through a \acrshort{MLP} and attended to the image features in our proposed \emph{Geometric Mapping} block, yielding feature-enhanced 3D points that form the \acrshort{NeMO}. A decoder attends a query image with the \acrshort{NeMO} using multiple Cross- and Self-Attention blocks to generate multiple dense predictions. We represent the \acrshort{NeMO} as a point cloud and reduce the higher dimensional \acrshort{NeMO} features to RGB using PCA~\cite{sarker2021ml}. The PCA reduction shows a relation between the learned features and the objects geometry and semantics.}\label{fig:method}
    \vspace{-1em}
\end{figure*}

\section{Related Work}\label{sec:related_work}
Our work focuses on the perception of previously \textit{unseen} objects, encompassing detection, segmentation, and pose estimation.
\textit{Unseen} refers to objects not explicitly seen during training, provided at runtime via 3D models (\textit{model-based}) or a set of RGB template images (\textit{model-free}).
While designed for the model-free setting, our method also supports model-based input. \\
\textbf{Unseen Object Segmentation and Detection.}
Given a target object, GFreeDet~\cite{liu2025gfreedet}, a model-free method, detects and segments objects through reconstructing a Gaussian object and comparing generated templates to region proposal from the query image.
NOCTIS~\cite{gandyra2025noctisnovelobjectcyclic} similarly computes segmentation masks by matching the appearance and semantic of reference images and a query image as obtained through Grounded SAM2 and DINOv2.\\
Instead of learning a presentation to render template images, model-based approaches directly use the provided CAD-model.
To this end, CNOS~\cite{nguyen_cnos_2023} matches DINOv2~\cite{oquab2023dinov2} tokens of template renderings with region proposals derived from SAM/ FastSAM~\cite{zhao2023fast, Kirillov_2023_ICCV}.
Similarly, NIDS-Net~\cite{lu2024adapting} also leverages SAM~\cite{Kirillov_2023_ICCV} and Grounding DINO~\cite{liu2024grounding} to obtain proposal embeddings which are compared to averaged and masked DINOv2's template embeddings of the object. OC-DiT~\cite{ulmer2025conditionallatentdiffusionmodels} uses latent diffusion models conditioned on template images to generate segmentation masks.\\
\textbf{Unseen Object Pose Estimation.}
Model-free pose estimation also commonly requires multiple reference registration images~\cite{sun_onepose_2022, park_latentfusion_2020, wen_foundationpose_2024, leonardis_grounding_2025}.
To this end, FS6D~\cite{he_fs6d_2022} predicts 6D poses by fusing features from support RGBD images and the query scene to establish dense correspondences.
OnePose~\cite{sun_onepose_2022} and OnePose++~\cite{he_oneposeplusplus_2022} employ pairwise image matching of dense 2D-3D correspondences hierarchically in a coarse-to-fine fashion.
RelPose~\cite{avidan_relpose_2022} estimates the relative camera rotation between image pairs through an energy-based formulation, which is extended to the relative 6D pose based on multiple images in the extension RelPose++~\cite{lin_relpose_2024}.
In contrast, PIZZA~\cite{nguyen_pizza_2022} approaches 6D object tracking through either an image pair or multiple template images between which the relative pose is estimated.
Similarly, BundleSDF~\cite{wen_bundlesdf_2023} jointly tracks poses and learns a Neural Object Field from a RGB-D stream, using RGB renderings from the learned field to supervise texture and geometry prediction.
FoundationPose~\cite{wen_foundationpose_2024} estimates either single-view poses for a model-based or a model-free scenario and also performs pose tracking.
The model-free setup is similar to BundleSDF, whereas in the model-based scenario templates are rendered based on the given model.\\
Analogously to detection and segmentation, model-based pose estimation relies on a CAD model of the unseen target object.
GigaPose~\cite{nguyen_gigapose_2024} relies on comparing and matching the most similar RGB template to the query image.
MegaPose~\cite{labbe_megapose_2023} applies a render-and-compare strategy to refine the best reference rendering of the target object.
Both methods rely on pre-training on a large-scale synthetic dataset spanning 2 million images.
OSOP~\cite{shugurov_osop_2022} establishes dense 2D-3D correspondences based on a template match, similar to ZS6D~\cite{ausserlechner_zs6d_2024}, which also relies on a self-supervised pre-trained \acrshort{ViT} for feature extraction.
SAM6D~\cite{Lin_2024_CVPR} predicts the 6D pose and the segmentation mask by applying a coarse-to-fine 3D-3D correspondence matching strategy that builds up on region proposal obtained from SAM and matched with template renderings \wrt~appearance, semantics and geometry.
Similarly, ZeroPose~\cite{zeropose_2025} also estimates the 6D pose and the segmentation masks given the CAD model and an RGB-D image as input. 
The method matchs the  query image with the feature embeddings derived from DINOv2 and prompted with the CAD model.

%% file: 3_method.tex
\section{Method}\label{sec:method}
\newcommand{\Fset}{\mathcal{F}}  %
\newcommand{\Iset}{\mathcal{I}}         %
\newcommand{\FsetI}{\Fset^{\Iset}} %
\newcommand{\Fsetq}{\Fset^{\text{q}}} %
\newcommand{\Qset}{\mathcal{Q}}         %
\newcommand{\FsetQ}{\Fset^{\Qset}} %
\newcommand{\Sset}{\mathcal{S}} %

\newcommand{\Felem}[1]{f_{#1}}          %
\newcommand{\Ielem}[1]{I_{#1}}          %
\newcommand{\FelemI}[1]{f^{\Iset}_{#1}}  %
\newcommand{\Felemq}[1]{f^{\text{q}}_{#1}}  %
\newcommand{\Qelem}[1]{q_{#1}}          %
\newcommand{\FelemQ}[1]{f^{\Qset}_{#1}}  %
\newcommand{\Selem}[1]{s_{#1}} %
\newcommand{\Selemgt}[1]{\Bar{s}_{#1}} %

\newcommand{\MLP}{\lambda} %
\newcommand{\UDF}{\mathcal{U}} %
\newcommand{\nemoencfunc}{\Psi}  %
\newcommand{\nemodecfunc}{\Theta} %

\newcommand{\nemosymbol}{\chi} %
\newcommand{\anchorImage}{\Ielem{\text{A}}}
\newcommand{\queryImage}{\Ielem{\text{q}}}

In the following, we present our encoder-decoder network as well as the proposed \acrfull{NeMO} representation.
\Cref{fig:method} gives an overview of the complete approach.
Our core idea is to separate visual and geometric object information from neural network weights, enabling an object-agnostic approach that adapts to any demonstrated target object without additional training.
To this end, given a set of RGB template images $\Iset = \left\{ \Ielem{i} \right\}_{i=1}^{K}$ of an object, where $K \geq 2$ and $\Ielem{i} \in \mathbb{R}^{H\times W\times 3}$ with $H$ and $W$ being the image height and width, we aim to construct a unified, object-centric representation without the need for intrinsic or extrinsic camera parameters.

\subsection{Network Architecture}\label{subsec:network_architecture}
\noindent\textbf{\gls{NeMO} Encoder}.
As Leap~\cite{jiang_leap_2023} has shown strong generalizability to novel objects, we use a similar attention mechanism in our network. A \gls{ViT}~\cite{dosovitskiy2020image} is used to extract patch-wise image features from all template images $\FsetI = \left\{ \FelemI{i} \right\}_{i=1}^{N}$ with $\FelemI{i} \in \mathbb{R}^{d}$ and $N=H_{\text{patch}}\times W_{\text{patch}}\times K$ where $H_{\text{patch}}, W_{\text{patch}}$ are the number of vertically and horizontally extracted patches respectively.
Since no prior information about the object coordinate frame or the camera-to-object transformation is provided, we define an anchor image $\anchorImage$.
It serves as the object's initial orientation in the \gls{NeMO} space.
During training, a random image from $\Iset$ acts as anchor $\anchorImage$.
As in~\cite{jiang_leap_2023}, we employ a multi-view encoder to generate updated image features $\widehat{\Fset}^{\Iset}$.
This part incorporates a strong bias towards the anchor image, using cross- and self-attention blocks that facilitate interactions between the anchor and non-anchor features.

\begin{figure}[t]
    \centering
    \includegraphics[width=1.0\columnwidth]{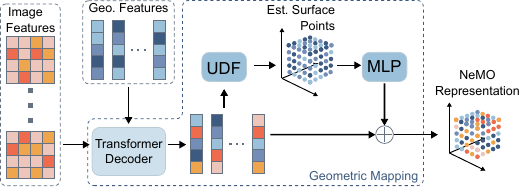}
    \caption{\textbf{Geometric Mapping Block.} We fuse the updated image features (key-value pairs) with the pre-processed geometric features (queries) in multiple transformer decoder blocks. The features are then forwarded to a \gls{UDF} that estimates the unsigned distance of the initial point cloud to the estimated object surface. After further processing these points via a \gls{MLP}, we combine them with the updated geometric features, resulting in the \gls{NeMO}.}\label{fig:nemo_geometric_mapping}
    \vspace{-1em}
\end{figure}

\noindent\textbf{Geometric Mapping.}
To enhance these features with the 3D geometry of the object oriented in the anchor coordinate system, we first create $\Qset = \left\{ \Qelem{i} \right\}_{i=1}^{M}$, a set of sampled 3D points $\Qelem{i} \in[-1, 1]^{3}$ with $M \geq 1$.
Next, we train a \gls{MLP} (see \suppref{fig:mlp_block}) to act as point encoder $\MLP$ that maps each 3D point $\Qelem{i}$ to a corresponding feature vector $\FsetQ = \left\{ \FelemQ{i} \right\}_{i=1}^{M} = \left\{ \MLP \left(\Qelem{i}\right) \right\}_{i=1}^{M}$ with $\FelemQ{i} \in \mathbb{R}^{d}$.
We fuse the resulting set of point features $\FsetQ$ and the updated image features $\widehat{\Fset}^{\Iset}$ via our proposed \emph{geometric mapping block}.
As shown in \cref{fig:nemo_geometric_mapping}, the initial point features $\FsetQ$ are updated with the information from $\widehat{\Fset}^{\Iset}$ through multiple Transformer Decoders:
\begin{equation}
    \begin{aligned}
        \widehat{\Fset}^{\Qset} = \left\{ \widehat{\Felem{i}}^{\Qset} \right\}_{i=1}^{M} = \text{TransformerDec}(\FsetQ, \widehat{\Fset}^{\Iset}) \in \mathbb{R}^{d},
    \end{aligned}
\end{equation}
where $\FsetQ$ acts as queries and $\widehat{\Fset}^{\Iset}$ as key-value pairs.
This allows the initial 3D point cloud features to attend to object-specific 2D image features.
Besides the visual cues, we also want to encode shape information of the object.
Therefore, we jointly learn a \gls{UDF} $\UDF$ using an \gls{MLP} that predicts the unsigned distance from each point $\Qelem{i}$ in the initial point cloud $\Qset$ to its closest point $\Selem{i} \in[-1, 1]^{3}$ on the estimated object surface $\Sset = \left\{ \Selem{i} \right\}_{i=1}^{M} $ \wrt the coordinate system of $\anchorImage$.
Note that $|\Qset| = |\Sset|$.
Since $\widehat{\Felem{i}}^{\Qset}$ depends on $\Qelem{i}$ we can define the distance $d_{i}$ and direction $v_{i}$ as
\begin{equation}
    \begin{aligned}
        d_{i} & =  \UDF\left(\widehat{\Felem{i}}^{\Qset}\left(\Qelem{i}\right)\right) \in \mathbb{R} \quad\mathrm{and}\quad
        v_{i} & = \frac{d\UDF\left(\widehat{\Felem{i}}^{\Qset}\left(\Qelem{i}\right)\right)}{d \Qelem{i}} \in \mathbb{R}^{\text{3}},
    \end{aligned}
\end{equation}
such that $\Selem{i} = \Qelem{i} - d_{i} v_{i}$. Note that $d_{i}$ is a scalar.

\noindent\textbf{\gls{NeMO} Representation.}
Given the $M$ estimated surface points from the set $\Sset$ and the processed 3D point features $\widehat{\Fset}^{\Qset}$, we define our \gls{NeMO} representation as $\nemosymbol = \left\{ \left( \Selem{i}, \widehat{\Felem{i}}^{\Qset} \right) \right\}_{i=1}^{M}$.
Keeping $\Selem{i}$ and $\widehat{\Felem{i}}^{\Qset}$ separate, we can transform the point cloud in \gls{NeMO} space, allowing us to free $\nemosymbol$ from the anchor coordinate system defined by $\anchorImage$.
This gives the advantage of being able to modify the geometric prediction of downstream tasks.
For convenience, we define the complete \gls{NeMO} encoder (see \cref{fig:method}) as a neural network $\nemoencfunc$ such that $\nemosymbol = \nemoencfunc(\Iset, \Qset)$.
To fuse the information of the predicted 3D points $\Selem{i}$ and their corresponding features we set $\widetilde{\Felem{i}}^{\Qset} = \widehat{\Felem{i}}^{\Qset} + \MLP(\Selem{i})$.
Our continuous \gls{UDF} approach differs from Leap~\cite{jiang_leap_2023}, which relies on a discrete, fixed-size neural volume.
We design our architecture to allow
\begin{enumerate*}[label=(\roman*)]
    \item variation in the number of points $M$, adjusting the size and descriptiveness of $\nemosymbol$,
    \item biased point sampling for $\Qset$, such as surface points from a CAD model,
    \item transformations of the \gls{NeMO} point cloud, allowing object modifications after the NeMO generation, and
    \item extension of an existing \gls{NeMO} with another set of \gls{NeMO} points.
\end{enumerate*}

\noindent\textbf{\gls{NeMO} Decoder.}
When observing a query image $\queryImage \notin \Iset$ the goal of our decoder $\nemodecfunc$ is to use the information stored in a \gls{NeMO}, which integrates information from all images in $\Iset$, to return a set of perception related dense predictions for the query image. As a first step the query image features $\Fsetq = \left\{\Felemq{i} \right\}_{i=1}^{H_{\text{patch}\times W_{\text{patch}}}}$ obtained by a \gls{ViT} are updated by the information stored in $\widetilde{\Felem{i}}^{\Qset}$ using a combination of cross- and self-attention layers as shown in~\cref{fig:method}. Inspired by the recent advancements in camera pose estimation and dense predictions~\cite{wang_dust3r_2024, leonardis_grounding_2025, wang2025vggt}, we use DPT~\cite{Ranftl_2021_ICCV} with multiple output heads to upscale the updated query image features $\widehat{\Fsetq}$ to multiple dense outputs $\nemodecfunc(\nemosymbol, \queryImage) = \left( P_{\text{modal}}, P_{\text{amodal}}, X, C \right)$, where $P_{\text{modal}} \in \mathbb{R}^{\text{H}\times\text{W}}$ and $P_{\text{amodal}} \in \mathbb{R}^{\text{H}\times\text{W}}$ are the predicted modal and amodal segmentation masks of the object, $X \in \mathbb{R}^{\text{H}\times\text{W}\times\text{3}}$ is the predicted dense pointmap between 2D pixels in $\queryImage$ and their corresponding 3D surface points in the coordinate system defined by $\nemosymbol$ and $C \in \mathbb{R}^{\text{H}\times\text{W}}$ is the associated learned confidence map to assess the 2D-3D mapping accuracy of $X$. After filtering the estimated  pointmap based on $C$, we utilize RANSAC~\cite{fischler_1981} and \gls{PnP}~\cite{Hartley_Zisserman_2004,lepetit_2009} to estimate the pose of the object in the query image.

\subsection{Training}
We train the encoder $\nemoencfunc$ and decoder $\nemodecfunc$ jointly end-to-end through multiple losses.
During training, we have access to a dataset of synthetically rendered RGB-D images with ground truth object poses, amodal and modal masks and camera intrinsic.
Given a training sample of multiple images of the same object, we randomly select one image as the anchor image $\anchorImage$ and one as a query image $\queryImage$ and use the ground truth poses together with the masked depth and camera intrinsic to build the ground truth surface points of the initial orientation of the \gls{NeMO} space.

\noindent\textbf{Losses.}
To enforce the encoder to predict towards ground truth surface points we define a simple regression loss that minimizes the Euclidean distance between the estimated surface point $\Selem{i} = \Qelem{i} - \widetilde{\UDF}(\Qelem{i}) \times \frac{d\widetilde{\UDF}(\Qelem{i})}{d\Qelem{i}}$ and the ground truth surface point $\Selemgt{i}$:
\begin{equation}
    \text{L}_{\nemosymbol}= \frac{\sum_{i=1}^{M} \lVert \Selem{i}-\Selemgt{i}\rVert}{M},
\end{equation}
with $M = |\nemosymbol|$.
We do not directly enforce any loss on the \gls{NeMO} features $\widehat{\Fset}^{\Qset}$ so that they can be freely learned through backpropagation through the decoder.
During training, after $\nemoencfunc$ has predicted $\nemosymbol$, we randomly rotate, translate and scale the \gls{NeMO} points $\Selem{i}$ by a random transformation $\mathbf{T}$ to teach the decoder to learn a coordinate system that is independent of the anchor image $\anchorImage$.
We define additional losses on the dense predictions of the decoder $\nemodecfunc$.
For the modal and amodal segmentation losses $\text{L}_{\text{modal}}$ and $\text{L}_{\text{amodal}}$ we use an equally weighted dice-loss~\cite{milletari_2016_dice_loss} and binary cross-entropy loss~\cite{jadon_2020}.
The pointmap loss $\text{L}_{\text{2D3D}}$ is a confidence weighted L1 loss which uses the estimated confidence map $C$ to weight the loss between the estimated 2D-3D correspondences $X$ and the ground truth mapping $\Bar{X}$. To learn the confidence map $C$, we define a certainty loss $\text{L}_{\text{certain}}$ and an uncertainty loss $\text{L}_{\text{uncertain}}$:
\begin{equation}
    \begin{aligned}
        \text{L}_{\text{certain}}   & = \frac{\sum\limits_{i \in \mathcal{D}_{\text{Obj}}} \Bigl( 1 - \tanh\bigl(\exp(C_{i})\bigr) \Bigr)}{\left| \mathcal{D}_{\text{Obj}} \right|} \\
        \quad \text{L}_{\text{uncertain}} & = \frac{\sum\limits_{i \in \mathcal{D}_{\text{Bg}}} \Bigl(\tanh\bigl(\exp(C_{i})\bigr)\Bigr)}{\left| \mathcal{D}_{\text{Bg}} \right|}
    \end{aligned}
\end{equation}
where $\mathcal{D}_{\text{Obj}}$ are the pixels belonging to the object and $\mathcal{D}_{\text{Bg}}$ are all pixels belonging to the background.
The total loss is a weighted sum between all losses. More details about the training in \suppref{subsec:losses}.

%% file: 4_experiment.tex
\section{Experiments}\label{sec:experiments}

\noindent\textbf{Synthetic Dataset.}
We create a new object-centric dataset using BlenderProc~\cite{Denninger2023} to generate \gls{PBR} images given the CAD object models as provided by a subset of Objaverse~\cite{objaverse}, GSO~\cite{9811809}, and OmniObject3D~\cite{wu2023omniobject3d}, resulting in a total of 11077 different objects.
We deem this necessary as there is -- to the best of our knowledge -- no available dataset of comparable object variety with sufficiently high but also balanced distribution of views per object.
For the following experiments, a single network is trained on parts of Objaverse and all OmniObject3D models. We supervised the training by evaluating on GSO objects, no fine-tuning or training on any of the objects present in the BOP benchmark is performed. For additional information we refer to \suppref{subsec:synthetic_data_generation}.

\noindent\textbf{Implementation.}\label{subsec:implementation}
We train our method on the aforementioned synthetic dataset and resize the respective, artificially corrupted object bounding box crops to $224 \times 224$.
We train for 400k steps (roughly 2000 epochs) with a maximum learning rate of $1\times10^{-4}$ on which we apply a linear warm-up of 5000 steps followed by standard cosine annealing.
The \gls{ViT} backbone is trained with a separate maximum learning rate of $1\times10^{-5}$. As \gls{ViT} we use DINOv2~\cite{oquab2023dinov2} and DPT~\cite{Ranftl_2021_ICCV} as regression head.
Optimization is performed using AdamW~\cite{loshchilov2018decoupled}.
Training takes roughly 10 days on 16 A100 GPUs with an effective batch size of 128, whereas the following experiments are run on a single A100~GPU.

\noindent\textbf{Experimental Setup.}\label{subsec:experimental_setup}
We evaluate our method's capability to perform multiple few-shot perception tasks in a model-free setting, \ie~no CAD model is given, and a model-based setting, \ie~a CAD model is given only during inference but not during training.
Following the BOP challenge's~\cite{hodan_bop_2024} dataset split we use the T-LESS~\cite{hodan2017tless}, TUD-L~\cite{Hodan_2018_ECCV} and YCB-V~\cite{xiang2018posecnn} datasets for model-based evaluation and the HOPEv2~\cite{tyree2022hope} and HANDAL~\cite{handaliros23} datasets for model-free evaluation. As metrics, we use \textit{Average Precision (AP)} and \textit{Average Recall (AR)} as defined in~\cite{hodan_bop_2024}.
In the model-free setup we use 32 randomly picked real RGB templates from the static onboarding videos provided by the BOP benchmark to generate the NeMO representation while in the model-based setting 32 \gls{PBR} rendered images are used.
The NeMO coordinate system is aligned with the ground truth object coordinate system as described in \suppref{subsec:object_alignment} for evaluation purposes only.
The dense decoder outputs are used for detection, segmentation and 6DoF pose estimation as described in \suppref{subsec:dense_to_pred}. Note that in both settings, the network weights are not changed, \ie no finetuning is performed and the objects have never been seen during training.
Additionally, we extensively ablate the NeMO representation and its influence on the downstream tasks in \cref{subsec:ablations} and show qualitative object surface reconstruction results of unknown objects in \cref{subsec:mf_reconstruction}. The same network is used for all experiments if not stated otherwise.

\subsection{Model-Free Few-Shot Perception}\label{subsec:mf_perception}
\begin{table}[t]
    \centering

    \resizebox{0.48\textwidth}{!}{
        \begin{tabular}{ l c c }
            \toprule
            Method                                                & HOPEv2            & HANDAL            \\
            \midrule
            CNOS (SAM) - Static onboarding~\cite{nguyen_cnos_2023} & 0.345             & --                \\
            dounseen-SAM-CTL~\cite{gouda2023dounseen}              & 0.380             & --                \\
            GFreeDet-FastSAM~\cite{liu2025gfreedet}                & 0.364             & 0.255             \\
            GFreeDet-SAM~\cite{liu2025gfreedet}                    & \underline{0.384} & \underline{0.264} \\
            Ours                                                  & \textbf{0.411}    & \textbf{0.273}    \\
            \bottomrule
        \end{tabular}
    }
    \caption{\textbf{Model-Free Detection.} We compare AP on \acrshort{BOP} test splits of HOPEv2 and HANDAL against other methods published on the public Model-Free Unseen Object 2D Detection leaderboard~\cite{leaderboard_model_free_det}.}\label{tab:model_free_det_results}
    \vspace{0.5em}
    \resizebox{0.48\textwidth}{!}{
        \begin{tabular}{ l l c c }
            \toprule
            Method                   & Detections                             & HOPEv2            & HANDAL            \\
            \midrule
            OPFormer$^{\text{\dag}}$ & CNOS~\cite{nguyen_cnos_2023}            & \textbf{0.335}    & 0.204             \\
            Ours                     & CNOS~\cite{nguyen_cnos_2023}            & 0.307             & --                \\
            Ours                     & GFreeDet-FastSAM~\cite{liu2025gfreedet} & \underline{0.329} & \underline{0.213} \\
            Ours                     & NeMO                                   & 0.302             & \textbf{0.235}    \\
            \bottomrule
        \end{tabular}
    }
    \caption{\textbf{Model-Free 6DoF Pose Estimation.}  We compare AP on \acrshort{BOP} test splits of HOPEv2 and HANDAL against other methods published on the public Model-Free Unseen Object 6D Detection leaderboard~\cite{leaderboard_model_free_poseest}. \dag~indicates unpublished methods.}\label{tab:model_free_poseest_results}
\vspace{-1em}
\end{table}
\begin{figure}
    \centering
    \includegraphics[width=0.95\linewidth]{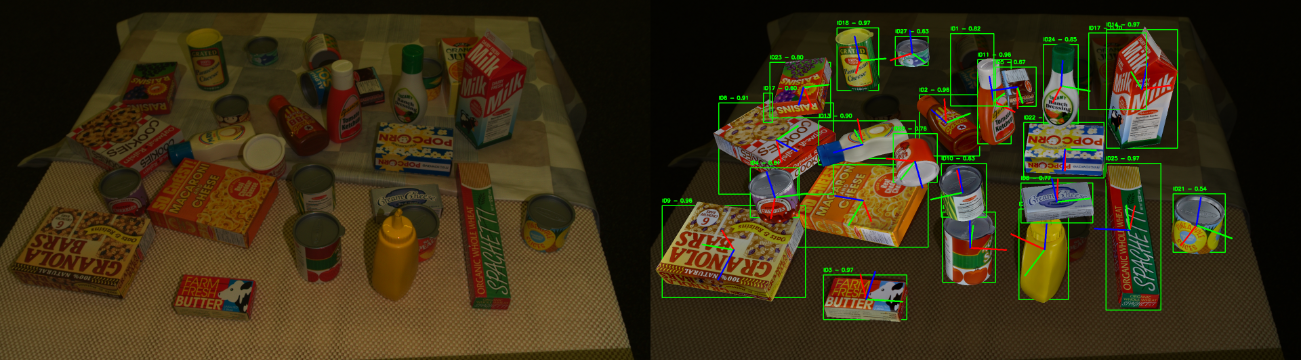}
    \caption{\textbf{Qualitative Example of Model-Free Few-Shot Detection and Pose Estimation on HOPEv2.} Left shows the scene without annotations, right shows \gls{NeMO} detections in green and pose estimations with refinement as rendered overlays. Even in the underexposed scene the model predicts reasonable results.}\label{fig:hopev2_preds_ICP}
    \vspace{-1em}
\end{figure}

\noindent\textbf{Model-Free Detection.}\label{subsec:mf_detection}
\cref{tab:model_free_det_results} shows the AP results of our model-free amodal detection on HOPEv2 and HANDAL datasets compared to all other publicly listed results. Our method achieves state-of-the art performance on both datasets, outperforming the previous best method by 2.7pp on HOPEv2 and 1.9pp on HANDAL.
While all other methods rely on SAM~\cite{Kirillov_2023_ICCV, zhao2023fast} segmentations of the scene for their bounding box predictions, we are, to the best of our knowledge, the first to use a single network to predict amodal segmentations/detections in a model-free setting. While SAM is able to give modal object segmentations, we can predict amodal segmentations based on the \gls{NeMO} representation, which we use to create amodal bounding boxes. Note that in the model-free category, the BOP benchmark does only evaluate amodal detection, no segmentation.

\noindent\textbf{Model-Free 6DoF Pose Estimation.}\label{subsec:mf_6dof}
We evaluate our model's capability for 6DoF Pose Estimation in a model-free setting on the HOPEv2 and HANDAL datasets using different detections in \cref{tab:model_free_poseest_results}. On HOPEv2 we use ICP~\cite{besl_1992_icp} between our estimated object surface and the depth information while no refinement is used on HANDAL, since no depth data is available. An example can be seen in \cref{fig:hopev2_preds_ICP}.
Compared to the only other method OPFormer we achieve state-of-the-art results on HANDAL when using \gls{NeMO} detections while being on par when using GFreeDet-FastSAM detections. When using the default CNOS detection as provided by the BOP benchmark, we are 2.8pp behind OPFormer. Note that as of the time of writing, no default CNOS detections for HANDAL are available anymore. Although our detections outperform previous methods on HOPEv2 they do not provide an AP gain for pose estimation. We hypothesize that the detection improvements come from our models ability to predict amodal bounding boxes, which is beneficial for the detection evaluation but might not always be an improvement for pose estimation.

\noindent\textbf{Model-Free Object Reconstruction.}\label{subsec:mf_reconstruction}
\begin{figure*}[t]
    \centering
    \includegraphics[width=0.9\linewidth]{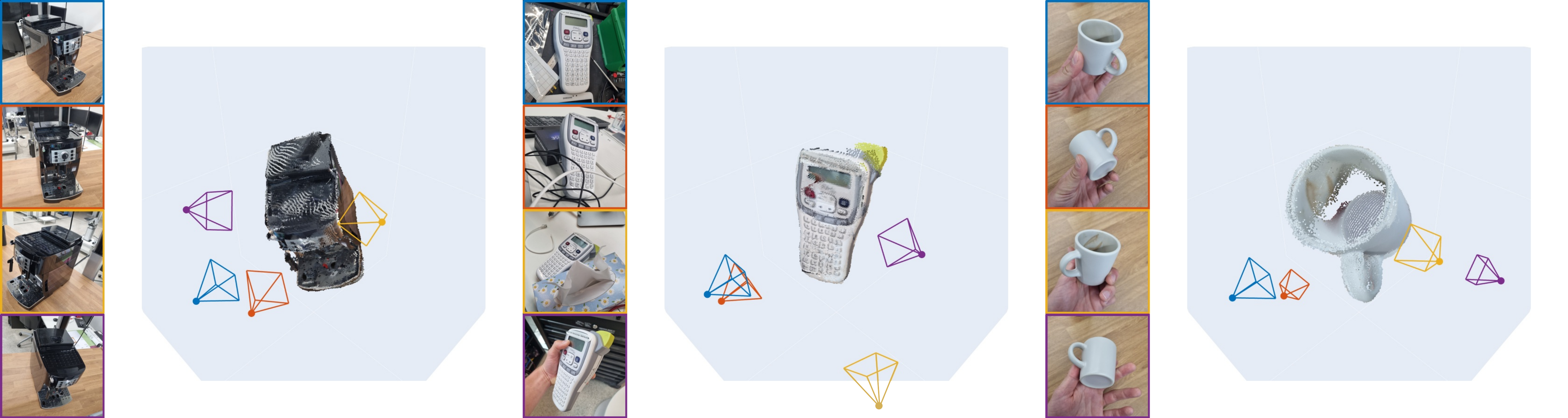}
    \caption{\textbf{Object Surface Reconstruction and Camera Pose Estimation on Unseen Objects.} We show object surface points and camera poses as predicted by the decoder based on four images of randomly chosen objects in different scenarios: (Left) A static coffee machine standing on a table, captured by a dynamic camera. (Middle) A label machine in different environments, including occlusions. (Right) An espresso mug manipulated in hand, captured by a static camera. In all three scenarios, our model is able to predict object-centric camera poses and surface points. We map RGB pixel color to corresponding 3D point to show correct 2D-3D mapping. Blue is the anchor image.}\label{fig:reconstruction_examples}
    \vspace{-1em}
\end{figure*}
We qualitatively show object surface reconstruction on random objects in different scenarios in \cref{fig:reconstruction_examples}.

\subsection{Model-Based Few-Shot Perception}\label{subsec:mb_perception}
\begin{table}[t]
    \centering
    \resizebox{0.48\textwidth}{!}{
        \begin{tabular}{ l c c c }
            \toprule
            Method                                & T-LESS            & TUD-L             & YCB-V             \\
            \midrule
            CNOS(Fast Sam)~\cite{nguyen_cnos_2023} & 0.395             & 0.534             & 0.568             \\
            SAM6D-FastSAM~\cite{Lin_2024_CVPR}     & 0.417             & 0.546             & 0.573             \\
            SAM6D~\cite{Lin_2024_CVPR}             & 0.458             & 0.573             & 0.589             \\
            F3Dt2D$^{\text{\dag}}$                & \underline{0.482} & 0.573             & 0.666             \\
            MUSE$^{\text{\dag}}$                  & 0.467             & 0.590             & \underline{0.674} \\
            anonymity$^{\text{\dag}}$             & 0.477             & \underline{0.593} & \textbf{0.685}    \\
            NIDS Net~\cite{lu2024adapting}         & \textbf{0.493}    & 0.486             & 0.621             \\
            Ours                                  & 0.183             & \textbf{0.623}    & 0.602             \\

            \bottomrule
        \end{tabular}
    }
    \caption{\textbf{Model-Based Detection.}  We compare AP on \acrshort{BOP} test splits of T-LESS, TUD-L and YCB-V against other methods published on the public Model-Based Unseen Object 2D Detection leaderboard~\cite{leaderboard_model_based_det}. \dag~indicates unpublished methods.}\label{tab:model_based_det_results}
    \vspace{-0.5em}
\end{table}

\begin{table}[t]
    \centering
    \resizebox{0.48\textwidth}{!}{
        \begin{tabular}{ l c c c }
            \toprule
            Method                                          & T-LESS            & TUD-L             & YCB-V             \\
            \midrule
            CNOS(Fast Sam)~\cite{nguyen_cnos_2023}           & 0.374             & 0.480             & 0.599             \\
            SAM6D-FastSAM~\cite{Lin_2024_CVPR}               & 0.420             & 0.517             & 0.621             \\
            SAM6D~\cite{Lin_2024_CVPR}                       & 0.451             & 0.569             & 0.605             \\
            NOCTIS~\cite{gandyra2025noctisnovelobjectcyclic} & 0.479             & 0.583             & \underline{0.684} \\
            LDSeg$^{\text{\dag}}$                           & \underline{0.488} & \underline{0.587} & 0.647             \\
            MUSE$^{\text{\dag}}$                            & 0.451             & 0.565             & 0.672             \\
            Prisma-MPG + SG$^{\text{\dag}}$                 & 0.454             & \textbf{0.590}    & 0.607             \\
            anonymity$^{\text{\dag}}$                       & 0.464             & 0.569             & \textbf{0.688}    \\
            NIDS Net~\cite{lu2024adapting}                   & \textbf{0.496}    & 0.556             & 0.650             \\
            Ours                                            & 0.169             & 0.488             & 0.579             \\

            \bottomrule
        \end{tabular}
    }
    \caption{\textbf{Model-Based Segmentation.} We compare AP on \acrshort{BOP} test splits of T-LESS, TUD-L and YCB-V against other methods published on the public Model-Based Unseen Object 2D Segmentation leaderboard~\cite{leaderboard_model_based_seg}. \dag~indicates unpublished methods.}\label{tab:model_based_seg_results}
    \vspace{-0.5em}
\end{table}

\begin{table}
    \centering
    \resizebox{0.48\textwidth}{!}{
        \begin{tabular}{ l l c c c }
            \toprule
            Method                              & Detections                  & T-LESS            & TUD-L             & YCB-V             \\
            \midrule
            Ours                                & NeMO                        & 0.082             & 0.466             & 0.493             \\
            Ours                                & ground truth                & 0.295             & 0.538             & 0.566             \\
            \hline
            ZS6D~\cite{ausserlechner_zs6d_2024}  & CNOS~\cite{nguyen_cnos_2023} & 0.210             & --                & 0.324             \\
            MegaPose~\cite{labbe_megapose_2023}  & CNOS~\cite{nguyen_cnos_2023} & 0.177             & 0.258             & 0.281             \\
            GenFlow~\cite{Moon_2024_CVPR}        & CNOS~\cite{nguyen_cnos_2023} & 0.215             & 0.300             & 0.277             \\
            GigaPose~\cite{nguyen_gigapose_2024} & CNOS~\cite{nguyen_cnos_2023} & 0.264             & 0.300             & 0.278             \\
            FoundPose~\cite{ornek2024foundpose}  & CNOS~\cite{nguyen_cnos_2023} & \underline{0.338} & 0.469             & 0.452             \\
            Co-op~\cite{Moon_2025_CVPR}          & CNOS~\cite{nguyen_cnos_2023} & \textbf{0.592}    & \textbf{0.642}    & \textbf{0.626}    \\
            Ours                                & CNOS~\cite{nguyen_cnos_2023} & 0.190             & \underline{0.476} & \underline{0.504} \\
            \bottomrule
        \end{tabular}
    }
    \caption{\textbf{Model-Based 6D Localization of Unseen Objects without Refinement.} We report \gls{AR} on 3 BOP datasets and compare with current SOTA model-based RGB based pose estimation methods \textit{without refinement}. We use the default CNOS~\cite{nguyen_cnos_2023} detections provided by the \acrshort{BOP} challenge when indicated. Data taken from~\cite{Moon_2025_CVPR}.}\label{tab:mb_6dof_results_no_refinement}
    \vspace{-1em}
\end{table}

This section discusses results of our network on model-based perception. Although the network has never been trained on rendered template images with black background and real query images, it performs on par and partially outperforms previous methods specialized on this task.

\noindent\textbf{Model-Based Detection.}\label{subsec:mb_detection}
AP for amodal detection on T-LESS, TUD-L and YCB-V datasets is reported in \cref{tab:model_based_det_results}. We achieve state-of-the-art results on TUD-L, outperforming the previous best method by 3pp. On YCB-V we would rank 6th out of 13 methods reported on the BOP leaderboard. On T-LESS we achieve a precision of 0.183, which is probably due to the strong similarities between the objects, the lack of texture as well as the dataset containing many cluttered scenes with objects of the same instance. All these factors are not present in our training data.

\noindent\textbf{Model-Based Segmentation.}\label{subsec:mb_segmentation}
In addition to detection we also report object segmentation AP on the three datasets. We report the results in \cref{tab:model_based_seg_results}. Compared to other networks our method is less precise on pixel level, which could be due to border artifacts as a results of patch-scaling. As with the detection results, the precision on T-LESS is low, which we attribute to the same reasons as mentioned above.

\noindent\textbf{Model-Based Refiner-Free 6DoF Pose Estimation.}\label{subsec:mb_6dof}
To evaluate the 6DoF Pose Estimation capabilities of our method independent of the detection quality we report the \textit{average recall AR} on T-LESS, TUD-L and YCB-V with default CNOS~\cite{nguyen_cnos_2023} detection and \textit{without additional refinement step} in \cref{tab:mb_6dof_results_no_refinement} as is standard practice in the literature.
Although our network was not designed for the model-based category we achieve high results in TUD-L and YCB-V, while only Co-op~\cite{Moon_2025_CVPR} achieves better results. Our method fails to handle the symmetric and textureless objects in T-LESS, which is reflected on the low average recall of 0.190. In addition we report the results of our method using ground truth and NeMO detections. Surprisingly, although the \gls{NeMO} detections show higher precision than the CNOS detections as reported in \cref{tab:model_based_det_results}, the average recall on pose estimation task is lower when using \gls{NeMO} detections. This is due to the difference in how average precision and average recall are evaluated.
For results on model-based pose estimation with refinement we refer to \suppref{tab:mb_pose_est_with_refinement}.

\subsection{Analyzing the \gls{NeMO} Representation}\label{subsec:ablations}
In this section we analyze the properties of the \gls{NeMO} representation. For all experiments we report AP of 6DoF pose estimation on the test split of YCB-V with real template images randomly chosen from the public real training set provided by the BOP benchmark~\cite{hodan_bop_2024}. We use ground truth detections and no refinement unless stated otherwise. A smaller decoder that only outputs pointmap and confidence is used in this section. Additional experiments and analyses can be found in \suppref{subsec:additional_nemo_analyses}.

\begin{figure}[t]
    \centering
    \includegraphics[width=1.0\columnwidth]{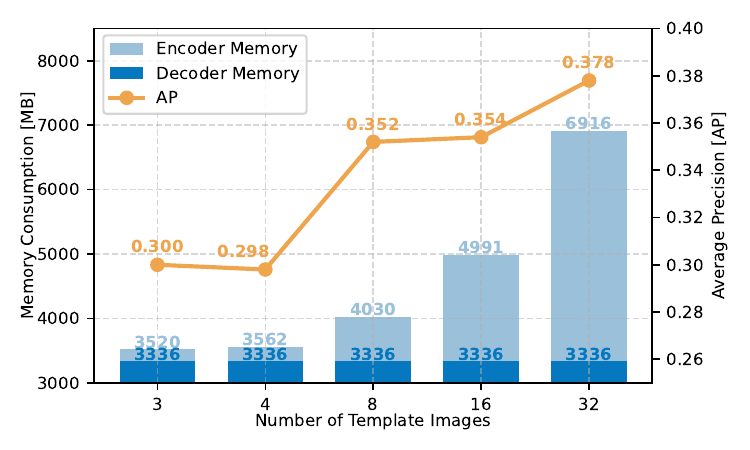}
    \vspace{-2em}
    \caption{\textbf{Memory consumption and AP on 6DoF Pose Estimation vs. Number of template images.}
        While the offline \gls{NeMO} generation requires more memory as the number of template images increases, the inference memory remains constant. Additionally, more template images increase the AP on 6DoF Pose Estimation on YCB-V with ground truth detections.}\label{fig:ap_mem_vs_num_templates}
        \vspace{-1em}
\end{figure}

\noindent\textbf{Varying number of template images.}
We show the relation between the number of template images used to generate a \gls{NeMO} and its influence on 6DoF pose estimation and the required memory footprint in \cref{fig:ap_mem_vs_num_templates}. It shows that while more template images enhance precision, acceptable performance is already achieved with just three views.
We emphasize the fact that whereas the number of templates increases, the runtime and memory consumption of our decoder model stays quasi constant while the precision increases.
Since we generate our \glspl{NeMO} before the inference task, we shift the computational heavy part of attending all template features with each other to the offline phase.

\begin{table}[t]
    \centering
    \resizebox{0.48\textwidth}{!}{
        \begin{tabular}{ c c c c c }
            \toprule
            \# NeMO Points & AP $\uparrow$ & AP$_{\text{MSPD}}$ $\uparrow$ & AP$_{\text{MSSD}}$ $\uparrow$ & Time/Image (s) $\downarrow$ \\
            \midrule
            10             & 0.004         & 0.005                         & 0.002                         & 0.402                       \\
            50             & 0.112         & 0.130                         & 0.094                         & 0.671                       \\
            100            & 0.214         & 0.231                         & 0.196                         & 0.460                       \\
            200            & 0.290         & 0.297                         & 0.282                         & 0.393                       \\
            500            & 0.380         & 0.379                         & 0.381                         & 0.340                       \\
            1000           & 0.383         & 0.389                         & 0.376                         & 0.320                       \\
            1500           & 0.378         & 0.37                          & 0.385                         & 0.336                       \\
            \bottomrule
        \end{tabular}
    }
    \caption{\textbf{Number of NeMO Input Points vs AP.} We report the AP on YCB-V 6D pose estimation with ground truth detections by varying number of randomly sampled input points.}\label{tab:num_nemo_points_vs_ap}
    \vspace{-1em}
\end{table}

\noindent\textbf{Varying number of NeMO Points.}
Although the model was trained on a fixed sized number of input points $\Qset$ we show in \cref{tab:num_nemo_points_vs_ap} that the encoder and decoder can adapt to different point cloud sizes, allowing for a dynamic adaptation of the models memory consumption and precision.
We observe that the precision increases as the number of \gls{NeMO} points increases up to 500 where it stagnates around 0.38. A visualization of \gls{NeMO} features can be seen in \cref{fig:method}.

\noindent\textbf{Transforming \gls{NeMO} coordinate system.}
During training, we randomly transform our \gls{NeMO} point cloud before passing it to the decoder.
In this section we evaluate if the decoder adapts its output based on the positions of the NeMO points.
To test the rotation-equivariance between the \gls{NeMO} point cloud and the predicted pointmaps $X$ of the template images, we rotate the \gls{NeMO} point cloud around the z-axis in 10 degree steps and evaluate the Chamfer distance~\cite{fan_cvpr_2017} between the predicted pointmap and the ground truth CAD model rotated by the same angle.
As can be seen in \suppref{fig:nemo_rot}, the Chamfer distance~\cite{fan_cvpr_2017} between the pointmaps and the ground truth CAD model remains low while the Chamfer distance between a non-rotating pointmap varies, indicating that the pointmap prediction is rotating accordingly.
This is an interesting property of our network for future work, in which parts of the \gls{NeMO} could be transformed online during inference.

\noindent\textbf{Extending NeMO.}
As the \gls{NeMO} representation is based on a set of points, it can easily be extended.
To see if the addition of new points from a different \gls{NeMO} of the same object leads to better results, we combine two \glspl{NeMO} with the same anchor image.
In \suppref{tab:extending_nemo} we show the Chamfer distance between the predicted pointmap and the ground truth CAD model for the original \gls{NeMO} and the extended one, observing that the Chamfer distance is lower for the extended \gls{NeMO} than for the original one.
This shows that we can combine two sets of \gls{NeMO} points without disturbing the decoder while improving its performance.
This is beneficial in scenarios, where the hardware is memory limited and thus, fewer template images can be used.

%% file: 5_conclusion.tex
\section{Limitations}\label{sec:limitations}
Despite promising results, our method exhibits several limitations. The encoder is trained to predict surface points, which leads to difficulties with symmetric objects, as demonstrated on the T-LESS dataset. We attribute this not to the pointmap representation itself, but to limitations in the current training procedure; addressing this will require further research. Additionally, the encoder performs poorly on highly textureless objects, likely due to their underrepresentation in the training set. Future work will focus on scaling the dataset to include a broader and more diverse set of objects. Another limitation is that the encoder does not directly predict bounding boxes; instead, segmentation masks are used as a proxy. This can result in merged bounding boxes when multiple instances of the same object are present—an issue we plan to resolve in future work.

\section{Conclusion}\label{sec:conclusion}
In this work, we presented an encoder and decoder architecture for \acrfull{NeMO}, a general and versatile object-centric representation that can be used for few-shot perception tasks such as object detection, segmentation and pose estimation using only an unordered set of RGB images. Through thorough experiments, we demonstrated that our method can be used for few-shot, model-free and model-based unseen perception task, partially outperforming state-of-the-art methods.
Our approach differs to alternative methods by
\begin{enumerate*}[label=(\roman*)]
    \item being capable to incorporate information from multiple RGB recorded images into a single representation without any camera parameters,
    \item utilizing a single network for multiple perception tasks,
    \item having constant inference time and memory requirements regardless of the number of template images,
    \item predicting amodal segmentation masks without CAD-model,
    \item being able to incorporate CAD-model information if given,
    \item allowing dynamically changing memory usage and precision, based on hardware capabilities,
    \item outsourcing the object's information from the network weights, allowing for quick adaptation to novel objects without retraining.
\end{enumerate*}
Furthermore, we contribute a realistic large-scale and balanced object-centric dataset that we deem beneficial for the broader research community.
Future work includes combining multiple \glspl{NeMO} for articulated objects.
Additionally, we strive to increase robustness of the applied task-specific encoder \wrt~symmetrical and textureless objects as motivated by the results on the T-LESS dataset.
We hope this work stimulates discussion on decoupling object knowledge from model weights and helps advance model-free few-shot perception for unseen objects.

%% file: 6_supplementary.tex
\clearpage
\section{Supplementary Material}\label{sec:supplementary}

\begin{figure}[h]
  \centering
  \includegraphics[width=0.3\columnwidth]{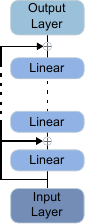}
  \caption{\textbf{\gls{MLP} Block with Skip-Connections.} }
  \label{fig:mlp_block}
\end{figure}

\subsection{Synthetic Training Data Generation}\label{subsec:synthetic_data_generation}
\begin{figure*}
  \centering
  \includegraphics[height=0.95\textheight]{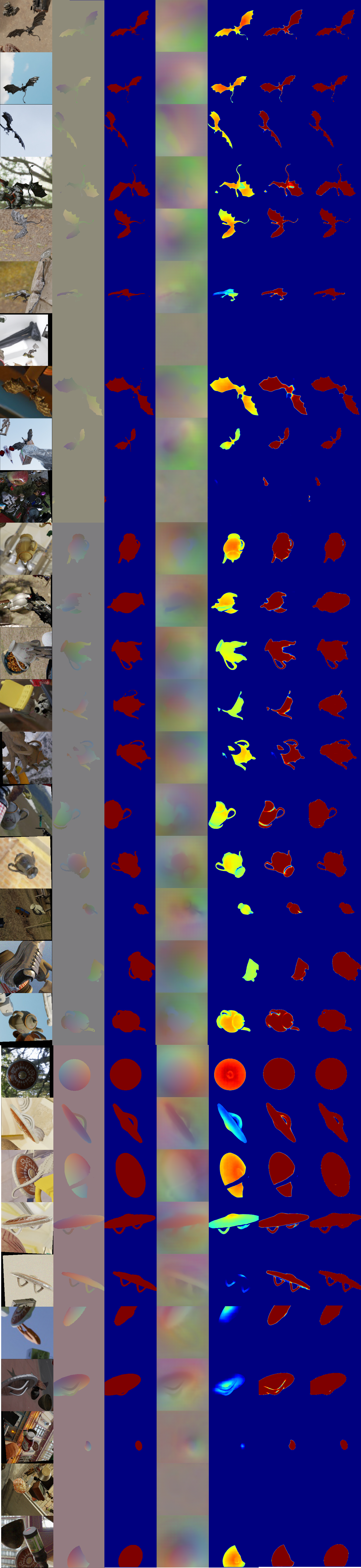}
  
  \caption{\textbf{Training Images.} From left to right, respectively: RGB input, ground truth 2D3D mapping, ground truth amodal segmentation, estimated 2D-3D mapping and estimated confidence, estimated modal segmentation and estimated amodal segmentation. Shown are three different training samples, each sample consists of 10 images. The first 5 images are used to generate a \gls{NeMO}, all 10 images are used for dense predictions. Note that in the last 5 templates of a sample, the object crop is stronger augmented and sometimes the object is not visible at all.}
  \label{fig:synt_imgs}
\end{figure*}

For the generation we proceed as following: We pick a random number of objects, sample them in a scene and render 20 \gls{PBR} images from random camera positions focusing on a random object in the scene for each camera position.
The scenes vary in objects placed
\begin{enumerate*}[label=(\roman*)]
    \item randomly on the ground,
    \item flying in space, or
    \item in an upright standing position.
\end{enumerate*}
We keep generating scenes until all objects have been seen (\ie at least 40\% visibility) at least 64 times in at least 8 different scenes.
Afterwards we convert the scene-centric into a highly efficient object-centric format by mapping the object id to all images where they occur, giving us at least 64 random views for each object in each scene type. See \cref{fig:synt_imgs} for some example images from the training set.

\subsection{Losses}\label{subsec:losses}
\vspace{0.3em}\noindent\textbf{Encoder loss.}
To enforce the encoder to predict towards ground truth surface points we define a simple regression loss that minimizes the Euclidean distance between the estimated surface point $\Selem{i} = \Qelem{i} - \widetilde{\UDF}(\Qelem{i}) \times \frac{d\widetilde{\UDF}(\Qelem{i})}{d\Qelem{i}}$ with $\widetilde{\UDF}(\Qelem{i}) := \UDF\left(\widehat{\Felem{i}}^{\Qset}\left(\Qelem{i}\right)\right)$ and the ground truth surface point $\Selemgt{i}$:
\begin{equation}
    \text{L}_{\nemosymbol}= \frac{\sum_{i=1}^{M} \lVert \Selem{i}-\Selemgt{i}\rVert}{M},
\end{equation}
with $M = |\nemosymbol|$.
We do not directly enforce any loss on the \gls{NeMO} features $\widehat{\Fset}^{\Qset}$ so that they can be freely learned through backpropagation through the decoder.
During training, after $\nemoencfunc$ has predicted $\nemosymbol$, we randomly rotate, translate and scale the \gls{NeMO} points $\Selem{i}$ by a random transformation $\mathbf{T}$ to teach the decoder to learn a coordinate system that is independent of the anchor image $\anchorImage$.

\vspace{0.3em}\noindent\textbf{Dense Prediction Losses.}
We define additional losses on the dense predictions of the decoder $\nemodecfunc$.
For the modal and amodal segmentation losses $\text{L}_{\text{modal}}$ and $\text{L}_{\text{amodal}}$ we use an equally weighted dice- and binary cross-entropy loss.
The pointmap is learned using a confidence weighted L1 loss:
We define the set of valid pixels in $\queryImage$ belonging to the object as $\mathcal{D}_{\text{Obj}}$.
As in~\cite{wang_dust3r_2024} we define a regression loss between the estimated 2D-3D correspondences $X$ and the ground truth mapping $\Bar{X}$:
\begin{equation}
    \text{L}_{\text{2D3D}} = \frac{\sum\limits_{i \in \mathcal{D}_{\text{Obj}}} \widetilde{C}_{i}\lVert X_{i} - \Bar{X}_{i} \rVert}{\mid \mathcal{D}_{\text{Obj}} \mid} \quad
\end{equation}
where $\widetilde{C}_{i}=1+\exp{C_{i}}$ makes sure that a confidence of zero does not lead to diminishing gradients.
Note that during training we transform the ground truth correspondences $\Bar{X}$ by the same transformation $\mathbf{T}$ as applied to the points $s_{i}$ in $\nemosymbol$.
Different to~\cite{wang_dust3r_2024}, we explicitly enforce certainty on pixels belonging to the object and uncertainty on pixels that do not belong to the object:
\begin{equation}
    \begin{aligned}
        \text{L}_{\text{certain}}   & = \frac{\sum\limits_{i \in \mathcal{D}_{\text{Obj}}} \Bigl( 1 - \tanh\bigl(\exp(C_{i})\bigr) \Bigr)}{\left| \mathcal{D}_{\text{Obj}} \right|} \\
        \quad \text{L}_{\text{uncertain}} & = \frac{\sum\limits_{i \in \mathcal{D}_{\text{Bg}}} \Bigl(\tanh\bigl(\exp(C_{i})\bigr)\Bigr)}{\left| \mathcal{D}_{\text{Bg}} \right|}
    \end{aligned}
\end{equation}
where $\mathcal{D}_{\text{Bg}}$ are all pixels not belonging to the object.
Based on the introduced losses, we define the total training loss as:
\begin{align}\label{eq:total_loss}
    \text{L}_{\text{total}} &= \alpha (\text{L}_{\nemosymbol} + \text{L}_{\text{2D3D}}) \nonumber \\
    &\quad + \beta (\text{L}_{\text{certain}} + \text{L}_{\text{uncertain}} + \text{L}_{\text{modal}} + \text{L}_{\text{amodal}}),
\end{align}
where $\alpha, \beta$ are hyper-parameters. During training we set $\alpha=1.0, \beta=0.2$.

\subsection{Training Details}\label{subsec:training_details}
\vspace{0.3em}\noindent\textbf{Sample Structure.} We define each sample in a batch as ten randomly picked images from the training set showing the same object and 1500 sampled points in the volume $[-1, 1]^{3}$. Five of the images are used as template images for the encoder to generate a \gls{NeMO}, while all ten images are used as input to the decoder. The total loss as defined in \cref{eq:total_loss} includes the encoder as well as multiple dense decoder losses, training the encoder and decoder jointly.
While the five images used for the \gls{NeMO} generation always show the same object with small variations in the bounding box, the other five images have a 30\% chance of being a completely random crop from the scene. This allows the decoder to learn to ignore images in which the object is not shown while also improving the decoder's capability to find the object in cluttered scenes.

\vspace{0.3em}\noindent\textbf{Data Augmentation.} In $2/3$ of the samples we use 1500 uniform sampled 3D points in the volume $[-1, 1]^{3}$ as input $\Qset$ while in $1/3$ of the samples we use 1500 points sampled close to the surface of the CAD model of the object. Due to the self-attention mechanism in the encoder, this allows us to train the model to use the CAD model information if available, while also being able to work without it.
As pixel-level data augmentation  to all templates in a sample, we apply:
\begin{itemize}
  \item 30\% chance of color jitter with 0.5 brightness, 0.5 contrast, 0.5 saturation and 0.5 hue, and
  \item 10\% chance of grayscale,
\end{itemize}
On template images directly we apply:
\begin{itemize}
  \item 50\% chance of random affine transformation with up to $5^\circ$ rotation, 0.1 translation and 0.1 scaling, and
  \item 50\% chance of Gaussian noise with standard deviation between 0 and 0.05.
\end{itemize}
\vspace{0.3em}\noindent\textbf{Initial \acrshort{NeMO} Space.}
To define the initial coordinate system based on the anchor image $\anchorImage$, we use the mean of the ground truth surface points as the center of the \gls{NeMO} space and the orientation of the anchor image as it is in the camera frame in OpenCV notation.
The scale of the \gls{NeMO} point cloud is defined such that the anchor image's surface points take $1/3$ of the volume $[-1, 1]^{3}$.

\subsection{Object Alignment}\label{subsec:object_alignment}
As the BOP challenge compares the predicted object pose against the ground truth pose, we need to align the coordinate system of our \gls{NeMO} with the ground truth object pose to be able to evaluate our approach. Note that this is only required for evaluation and generally not a necessity of our method. As we are given the object-to-camera transformation both in the model-based, where we render our templates, and in the model-free setting, where we are given the ground truth pose, we can align our coordinate system by optimizing the scale, rotation and center-offset. We use a simple gradient descent optimization that minimizes the error between the predicted and ground truth pose of the $\text{k}=5$ best rotation predictions of the template images. See \cref{alg:align} for the pseudo code of the alignment algorithm.

\subsection{From Dense Predictions to Detection, Segmentation and 6DoF Pose Estimation}\label{subsec:dense_to_pred}
Since our decoder produces dense predictions, we need additional steps to get a 6DoF pose estimation and amodal bounding box. These predictions are run in parallel using multiple decoder and different NeMOs. To predict the segmentations and amodal bounding boxes of a scene, we patchify the scene image with two patch sizes (smallest image side and half the smallest image side) with a stride of $1/3$ of the patch size. We accumulate the patch predictions and look for clusters of predictions larger 0 to differentiate between object instances. The amodal bounding box is retrieved by laying a box around a amodal segmentation cluster. We ignore small clusters and clusters touching the image sides. To predict the 6DoF pose estimation, we crop the region from a bounding box and all pixels with a minimum confidence value of 0.1 in the 2D-3D correspondences. We then compute the pose by solving the PnP problem using OpenCV's \texttt{solvePnPRansac} function with \texttt{iterationsCount 500}, \texttt{reprojectionError 6} and \texttt{min inlier ratio} of 30\%.

\begin{algorithm}[t]
\caption{NeMO Alignment Algorithm}
\label{alg:align}
\begin{algorithmic}[1]
\Require Ground truth poses $T^\text{gt}_i \in SE(3)$, estimated poses $T^\text{est}_i \in SE(3)$, max iterations $N$, learning rate $\eta$, Huber loss threshold $\delta$, number of best poses $k$
\Ensure Optimal scale $s$, rotation $R$, translation $t$ that aligns $T^\text{est}$ to $T^\text{gt}$

\State Select $k$ best estimates with lowest initial rotation error
\State Initialize scale $s$ based on average translation norms
\State Initialize rotation as axis-angle vector $\mathbf{r} \gets 0$, translation $\mathbf{t} \gets 0$
\For{$i = 1$ to $N$}
    \State $R \gets$ \Call{AxisAngleToRotation}{$\mathbf{r}$}
    \State Construct correction transform $T^\text{corr} = \begin{bmatrix} R & \mathbf{t} \\ 0 & 1 \end{bmatrix}$
    \State Apply correction: $\hat{T}_i \gets T^\text{est}_i \cdot T^\text{corr}$
    \State Compute predicted $\hat{R}_i$ and scaled $\hat{t}_i$
    \State Compute rotation error $\theta_i = \angle(\hat{R}_i, R^\text{gt}_i)$
    \State Compute translation error $\epsilon_i = \lVert \hat{t}_i - t^\text{gt}_i \rVert$
    \State Compute robust loss using Huber: $L = \text{Huber}(\theta) + \text{Huber}(\epsilon / 10)$
    \State Update $s$, $\mathbf{r}$, $\mathbf{t}$ via gradient descent with Adam optimizer
    \If{$L < L_\text{best}$}
        \State Save current $s$, $R$, $\mathbf{t}$ as best parameters
    \EndIf
\EndFor
\State \Return Optimized $s$, $R$, and $t$
\end{algorithmic}
\end{algorithm}

\subsection{Additional \gls{NeMO} Analyses}\label{subsec:additional_nemo_analyses}
\vspace{0.3em}\noindent\textbf{Similarity between predicted and ground truth object surface.} We show the close resemblance of our dense point cloud by measuring the Chamfer distance to the ground truth point cloud.
The prediction is based on two different point cloud sampling methods as input to our \gls{NeMO}.
Both methods achieve strong results in \cref{tab:cd_pc_sampling} across the three datasets with slightly better performance, when the input point cloud is sampled based on the actual CAD model for a given object.
\begin{table}[h]
  \centering
  \begin{tabular}{ c c c c }
    \toprule
    Sampling Method & \multicolumn{3}{c}{Chamfer Distance $\downarrow$}                   \\
                    & YCB-V                                             & HOPEv2 & HANDAL \\
    \midrule
    Random Sampled  & 0.0055                                            & 0.0029 & 0.009  \\
    CAD Surface     & 0.0036                                            & 0.0019 & 0.0022 \\
    \bottomrule
  \end{tabular}
  \caption{\textbf{Chamfer distance between ground truth point cloud and dense point cloud prediction based on two different \gls{NeMO} input sample point clouds.}
    First, randomly sampled points.
    Seconds, points sampled on the CAD surface, which requires the model of a given object.
    We observe already small errors only in case of randomly sampled point clouds as input to our \gls{NeMO} representation.
    We further improve this distance when we sample points based on the provided CAD model.}
  \label{tab:cd_pc_sampling}
\end{table}

\begin{figure}
    \centering
    \begin{subfigure}{0.49\columnwidth}
        \includegraphics[width=\linewidth]{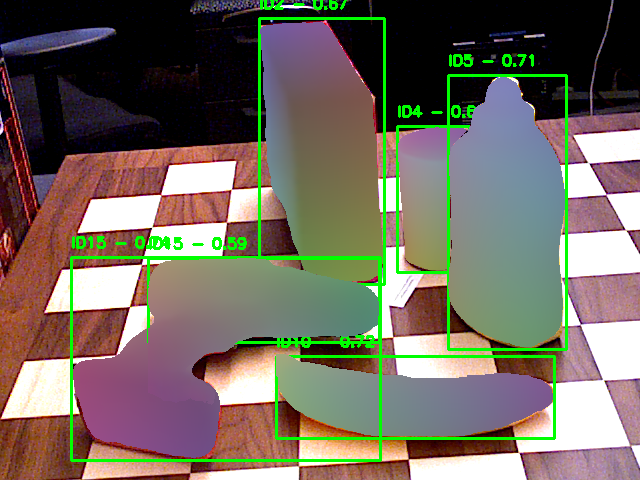}
    \end{subfigure}
    \hfill
    \begin{subfigure}{0.49\columnwidth}
        \includegraphics[width=\linewidth]{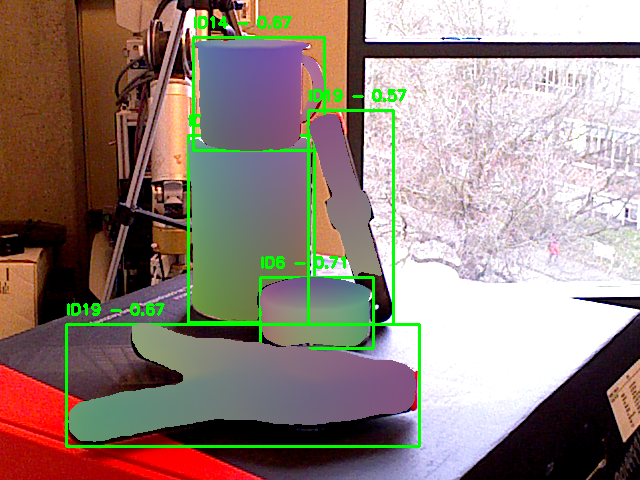}
    \end{subfigure}
    \hfill
    \caption{\textbf{Pointmap predictions} on the YCB-V dataset. We only show estimations with confidences $> 0.1$.
    }
    \label{fig:2d_3d_conf_ycbv}
\end{figure}

\noindent\textbf{CAD-Model Conditioned Sampling.}
We compare the performance of our network using randomly sampled $\Qset$ and points sampled from the surface of a CAD model. Additionally, we compare different ICP-based refinement strategies: \begin{enumerate*}[label=(\roman*)]
    \item no ICP,
    \item using the pointmap prediction as source point cloud for the ICP, and
    \item using CAD model surface points as source point cloud for the ICP.
\end{enumerate*}
The results are shown in \cref{tab:icp_pointsampling_ablation}. We show that our design choice to use a continuous point cloud as input to the encoder can leverage the information from the CAD model to improve the performance of our method. We also show that we can use the pointmap prediction for ICP refinement, although no depth information is given during the encoding step, increasing the average precision compared to purely RGB based (no ICP) pose estimation. An example of estimated object surfaces can be seen in \cref{fig:2d_3d_conf_ycbv}. This experiment demonstrates the versatility of our network, allowing us to adapt to different scenarios.
\begin{table}[t]
    \centering
    \resizebox{0.48\textwidth}{!}{
        \begin{tabular}{ l l c c c c }
            \toprule
            \acrshort{ICP} Method & Point Sampling & AP $\uparrow$ & AP$_\text{MSPD}$ $\uparrow$ & AP$_\text{MSSD}$ $\uparrow$ & Time/Image (s) $\downarrow$ \\
            \midrule
            --                    & Random         & 0.378         & 0.370                       & 0.385                       & 0.336                       \\
            --                    & CAD Surface    & 0.383         & 0.420                       & 0.347                       & 0.531                       \\
            Pointmap Surface      & Random         & 0.475         & 0.394                       & 0.557                       & 1.720                       \\
            Pointmap Surface      & CAD Surface    & 0.507         & 0.447                       & 0.567                       & 1.479                       \\
            CAD                   & Random         & 0.671         & 0.639                       & 0.703                       & 2.049                       \\
            CAD                   & CAD Surface    & 0.626         & 0.594                       & 0.658                       & 1.607                       \\
            \bottomrule
        \end{tabular}
    }
    \caption{\textbf{Precision with different point sampling and \acrshort{ICP}.} We report the AP on YCB-V with ground truth detections without refinement.}\label{tab:icp_pointsampling_ablation}
\end{table}

\begin{table}[t]
    \centering
    \resizebox{0.48\textwidth}{!}{
        \begin{tabular}{ l l c c c }
            \toprule
            Method                                                & T-LESS            & TUD-L             & YCB-V             \\
            \midrule
            Co-op (F3DT2D, Coarse, RGBD)\cite{Moon_2025_CVPR} & 0.620 & 0.841 & 0.808 \\
            FreeZeV2.1\cite{caraffa2024freeze}& 0.751 & 0.991 & 0.905 \\
            FreeZeV2.2\cite{caraffa2024freeze} & \underline{0.777} & \underline{0.988} & \underline{0.911} \\
            FRTPose-WAPP (Default Detection) &  \textbf{0.826} & 0.978 & \textbf{0.918}  \\
            Ours (CNOS Detection) & 0.291 & 0.560 & 0.592 \\
            Ours (NeMO detection) & 0.144 & 0.756 & 0.567 \\
            \bottomrule
        \end{tabular}
    }
    \caption{\textbf{Model-Based 6D Detection of Unseen Objects with Refinement.} We report \gls{AP} on 3 BOP datasets and compare against other methods published on the public Model-Based Unseen Object 6D Detection leaderboard.~\cite{leaderboard_model_based_poseest}. We used a simple ICP as refinement step. \dag~indicates unpublished methods.}\label{tab:mb_pose_est_with_refinement}
\end{table}

\begin{figure*}
  \centering
  \includegraphics[height=0.95\textheight]{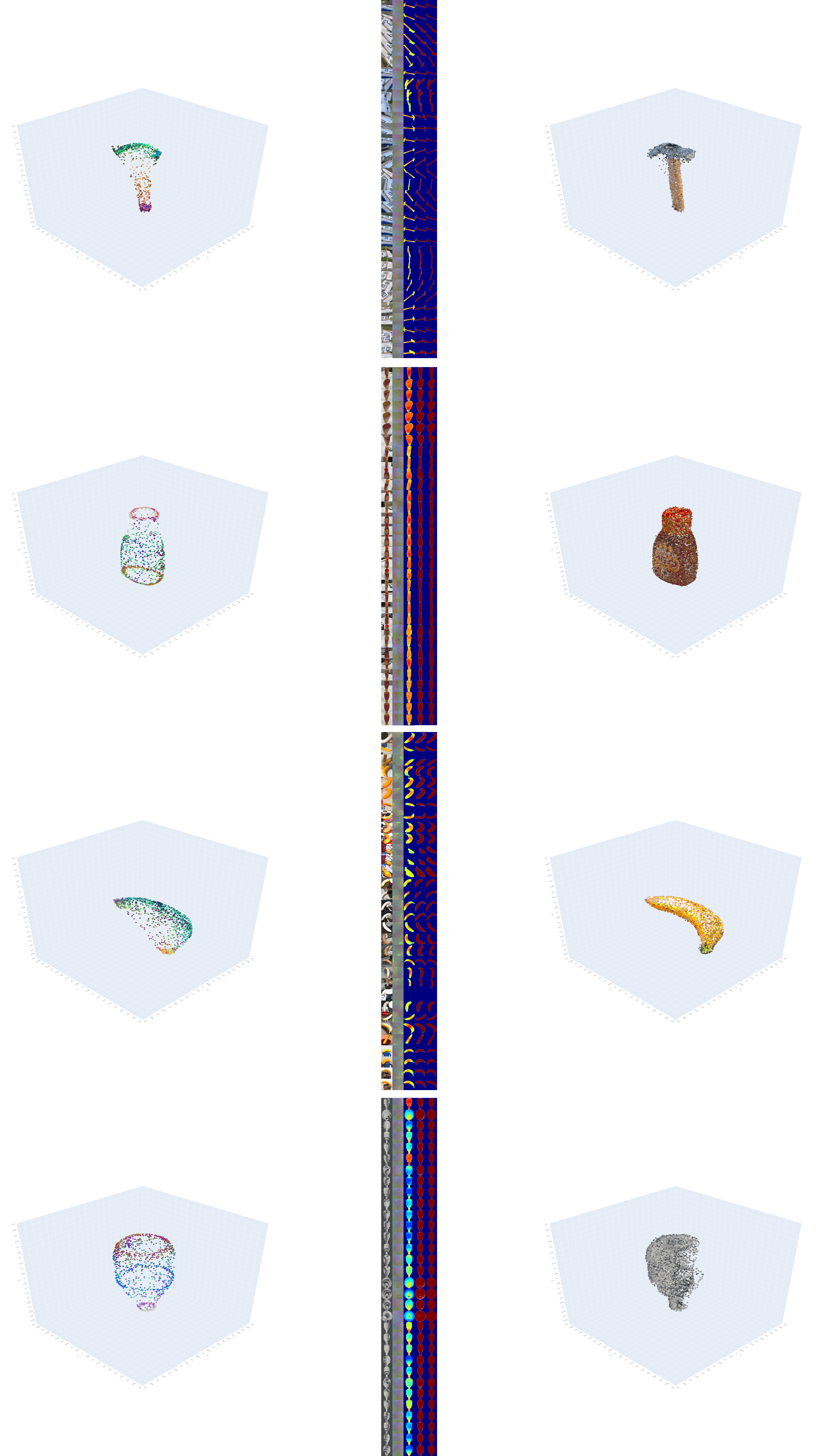}
	\caption{\textbf{NeMOs from benchmark objects.} We show the NeMOs (left, with PCA feature visualization), dense predictions (middle) and object surface reconstruction (right) from objects selected from different datasets. From top to bottom, the datasets are: HANDAL, HOPEv2, YCB-V and T-LESS. We use 32 template images to generate the NeMOs and use the same 32 images to show the dense prediction results. We show real template images for HANDAL, HOPEv2 and YCB-V and synthetic template images for the T-LESS object. NeMOs are aligned with ground truth poses as described in \cref{subsec:object_alignment}.}
  \label{fig:dataset_nemos}
\end{figure*}

\begin{figure}[t]
    \centering
    \includegraphics[width=\linewidth]{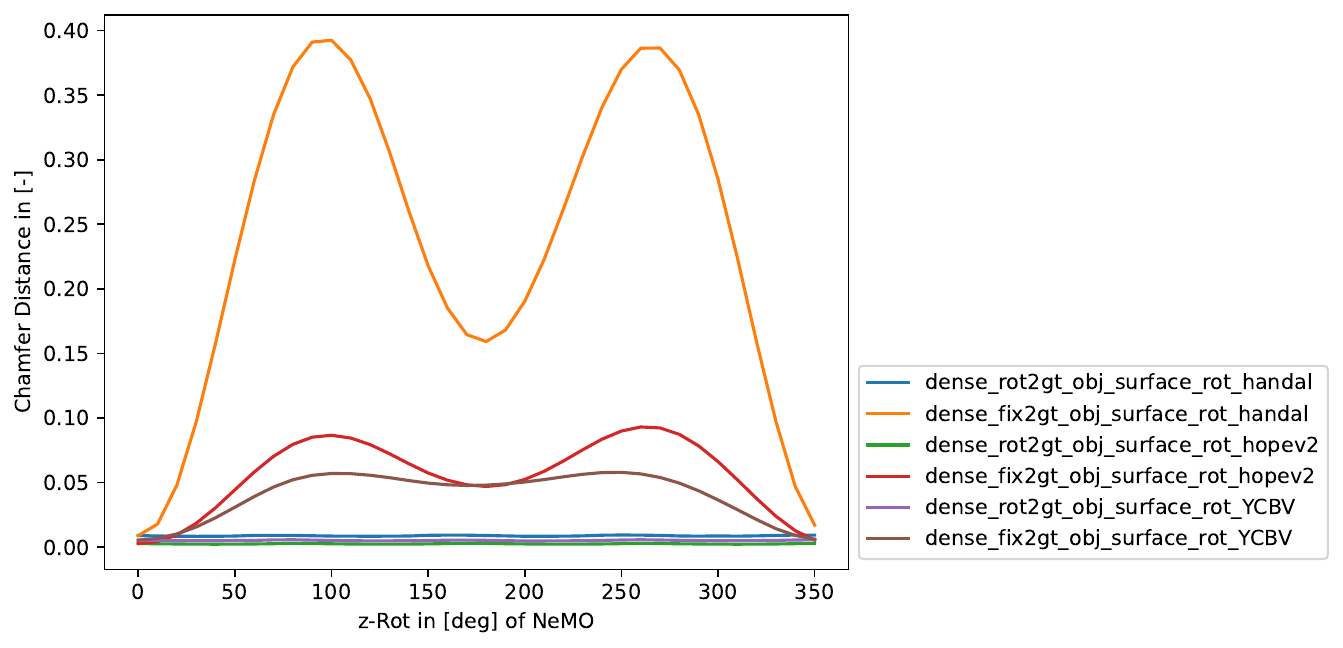}
    \caption{\textbf{Chamfer distance between dense predicted and rotated ground truth point cloud} averaged across all objects of respective dataset.
        In case objects and \glspl{NeMO} are rotated analogously, we observe small errors \ie the point clouds are very similar.
        M-shaped error trajectory, with minimum around 180 degrees of rotation around objects' z-axis, in case of fixed 2D-3D dense representation but rotated ground truth object point cloud.
        This shows that our representation is aware of rotations.}
    \vspace{-1em}\label{fig:nemo_rot}
\end{figure}

\begin{table}[t]
    \centering
    \begin{tabular}{ c c c c }
        \toprule
        \gls{NeMO} Generation & \multicolumn{3}{c}{Chamfer Distance $\downarrow$}                   \\
                              & YCB-V                                             & HOPEv2 & HANDAL \\
        \midrule
        Fused                 & 0.0080                                            & 0.0034 & 0.0088 \\
        All Templates         & 0.0081                                            & 0.0034 & 0.0084 \\
        \bottomrule
    \end{tabular}
    \caption{\textbf{Chamfer distance between ground truth point cloud and dense point cloud prediction based on two different \glspl{NeMO}.}
        First, a fused \gls{NeMO} based on two respective \glspl{NeMO} generated through 4 images each but each with the same anchor image.
        Second, predicted dense point cloud based on all 7 template views.
        This shows the versatility of our \gls{NeMO} as the point cloud resemblance and error stays constant.
    }\label{tab:extending_nemo}
    \vspace{-1em}
\end{table}